\documentclass[a4paper,num-refs]{oup-contemporary}

\journal{gigascience}

\usepackage{graphicx}
\usepackage{siunitx}

\usepackage{multirow}

\title{A dataset of ant colonies motion trajectories in indoor and outdoor scenes for social cluster behavior study}

\author[1,\authfn{2}]{Meihong Wu}
\author[1,\authfn{2}]{Xiaoyan Cao}
\author[2]{Xiaoyu Cao}
\author[1,\authfn{1}]{Shihui Guo}

\affil[1]{School of Informatics, Xiamen University, Xiamen, 361000, China}
\affil[2]{Chemistry and Chemical Engineering, Xiamen University, Xiamen, 361000, China}

\authnote{\authfn{1}guoshihui@xmu.edu.cn}
\authnote{\authfn{2}Contributed equally.}

\papercat{Paper}

\runningauthor{Meihong Wu et al.}

\jvolume{00}
\jnumber{0}
\jyear{2017}

\begin{document}

\begin{frontmatter}
\maketitle
\begin{abstract}
Motion and interaction of social insects (such as ants) have been studied by many researchers to understand the clustering mechanism. Most studies in the field of ant behavior have only focused on indoor environments, while outdoor environments are still underexplored. In this paper, we collect 10 videos of ant colonies from different indoor and outdoor scenes. And we develop an image sequence marking software named VisualMarkData, which enables us to provide annotations of ants in the video. In all 5354 frames, the location information and the identification number of each ant are recorded for a total of 712 ants and 114112 annotations. Moreover, we provide visual analysis tools to assess and validate the technical quality and reproducibility of our data. It is hoped that this dataset will contribute to a deeper exploration on the behavior of the ant colony.
\end{abstract}

\end{frontmatter}

\section{Background}
\label{sec:background}


Social insects often tend to cluster into a colony~\cite{vandermeer2008clusters}, which forms a complex social network~\cite{balch2001automatically}. 
From time to time, the social network springs up with self-organized clustering behaviors, including division of labor~\cite{holldobler1990ants}, task specialization~\cite{whitehouse1996ant}, and distributed problem solving~\cite{vaughan2000whistling}.
Biologists analyze the evolution of social network to understand the clustering behavior of insects~\cite{fewell2003social}, thus promoting the development of relevant modern applications, such as wireless communication~\cite{motani2005peoplenet} and cluster intelligent control~\cite{tiacharoen2012design}.
The key requirement of this research is the ability to track the motions and interactions of each individual robustly and accurately.

Until the late $20^{th}$ century, biologists still manually marked the motion trajectories on the video to guarantee the quality. However, they have to track each individual at one time, which might mean watching the entire video 50 times or more in a crowded scene.~\cite{poff2012efficient}
Obviously, manually tracking is time-consuming and prone to human error.
It becomes an inhibiting factor in obtaining the complete and accurate dataset required to analyze the evolution of social networks.
Therefore, in the past two decades, attempts have been made to automate the tracking process for social insects utilizing computer vision (CV) techniques~\cite{khan2005mcmc, khan2006mcmc, oh2006parameterized, veeraraghavan2008shape, fletcher2011multiple}.

Traditional CV techniques free researchers from manual work through approaches such as foreground segmentation algorithm~\cite{li2008estimating}, temporal difference method~\cite{khan2005mcmc} and hungarian algorithm~\cite{li2009learning}.
Such approaches, however, have failed to address the noise in the image~\cite{zhao2015improved}, resulting in the limitation that a laboratory environment with a clean background is needed.
Nevertheless, many scientifically valuable results are obtained in nature rather than laboratory environment~\cite{schmelzer2009special, kastberger2013social, tan2016honey, dong2018olfactory}.

Fortunately, with the emergence of deep learning, CV techniques are already capable of addressing many complex tasks~\cite{schor2016robotic, wang2018interactive,wang2018research}, which brings a piece of good news to automated insect tracking in outdoor scenes.
Several studies have explored automated multi-ant tracking in outdoor scenes using deep learning-based models~\cite{imirzian2019automated, cao2020online}.
Experimental results demonstrate that these models could be scaled up into a cost-effective alternative to traditional manual tracking methods which are typically costly and/or labor intensive~\cite{imirzian2019automated, cao2020online}.
A critical requirement for the development of these models is access to the datasets containing motion trajectories of insects in the video.
To the best of our knowledge, however, the current studies are all using only one outdoor-scene dataset, which is lack of data diversity.

In this paper, we collect video recordings of 10 ant colonies from different scenes (including indoor and outdoor scenes). 
Besides, we develop an image sequence mark software named VisualMarkData, which is used to mark the pixel patches covered by ants in each frame of the video.
The total size of our dataset is 5354 frames, 712 ants, and 114112 labels.
We believe that our dataset will benefit future research on social insect behavior analysis.


\begin{figure*}[ht]
\centering
\includegraphics[width=\linewidth]{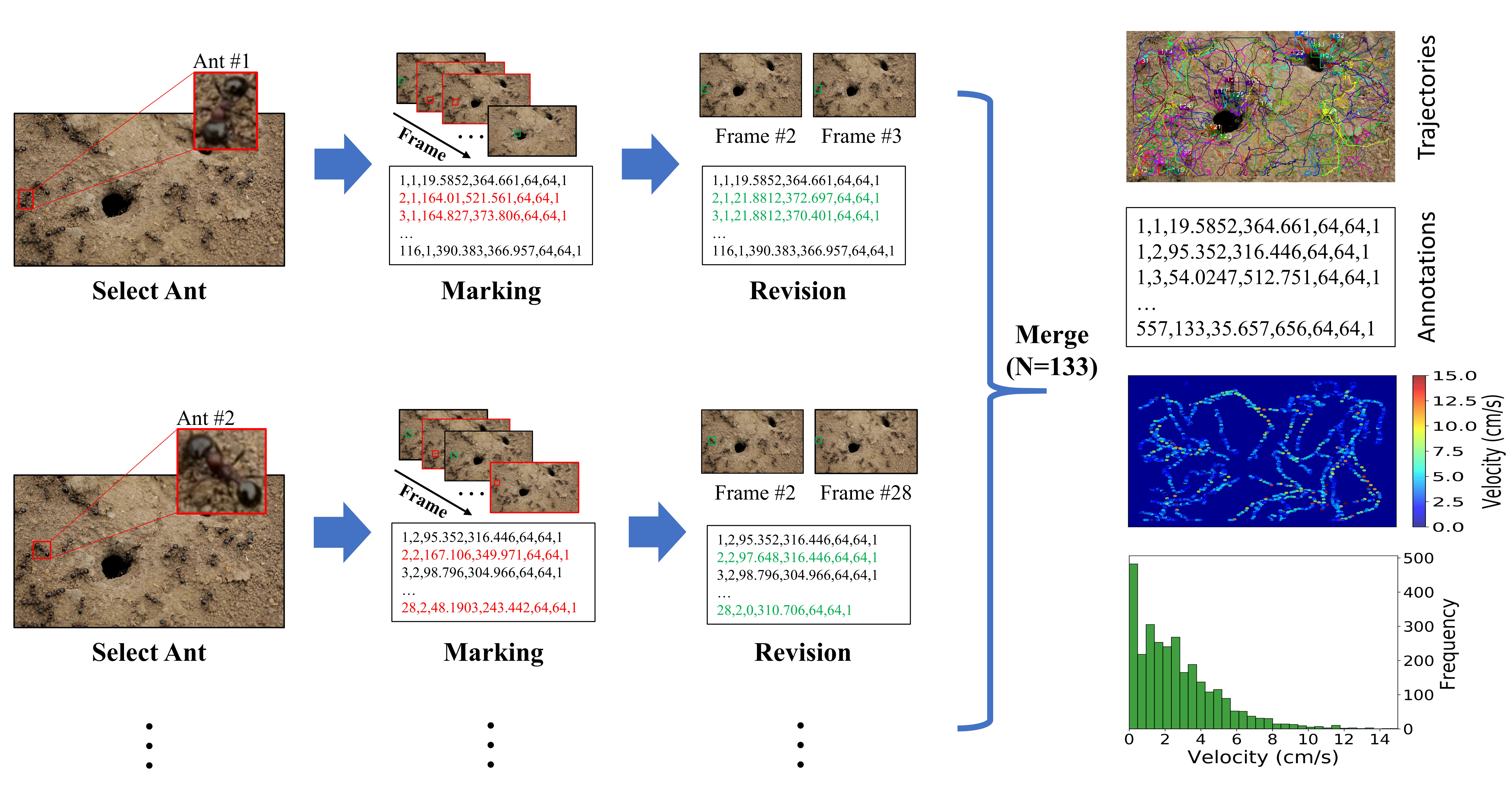}
\caption{The pipeline for marking motion trajectories of ants in an image sequence, taking an outdoor scene as an example.
A total of 133 ants appear in this image sequence, and we select one ant to be marked in each epoch.
We use a square bounding box to point out the ant's location and record the relevant parameters at the same time.
After all ants of the entire image sequence have been marked, we check the quality of the annotations frame by frame so that wrong annotations (red font) can be corrected (green font).
Then, we merge all the annotations of the image sequence into one file.
Additionally, we provide three visualization tools to verify the data quality.}
\label{fig:pipeline}
\end{figure*}

\section{Data Description}


We collect 10 videos that record activities of different ant colonies, including colonies from both indoor and outdoor scenes.
To help us mark the motion trajectories, we develop an image sequence marking software called VisualMarkData.

After spending a large quantity of time and effort, we obtain a dataset with 5354 frames and 114112 annotations.
Table~\ref{tab:Statistics} shows the statistical information of our dataset.

\begin{table*}[tbhp]
\centering
\begin{tabular}{|c|c|c|c|c|c|}
\hline
\textbf{Sequence} & \textbf{FPS}        & \textbf{Resolution}        & \textbf{Length} & \textbf{Ants} & \textbf{Annotations} \\ \hline
Seq0001           & \multirow{5}{*}{25} & \multirow{5}{*}{1920$\times$1080} & 351 (00:14)     & 10              & 3510           \\ \cline{1-1} \cline{4-6} 
Seq0002           &                     &                            & 351 (00:14)     & 10              & 3510           \\ \cline{1-1} \cline{4-6} 
Seq0003           &                     &                            & 351 (00:14)     & 10              & 3510           \\ \cline{1-1} \cline{4-6} 
Seq0004           &                     &                            & 351 (00:14)     & 10              & 3510           \\ \cline{1-1} \cline{4-6} 
Seq0005           &                     &                            & 1001 (00:40)    & 10              & 3510           \\ \hline
Seq0006           & \multirow{5}{*}{30} & \multirow{5}{*}{1280$\times$720}  & 600 (00:20)     & 73              & 11178          \\ \cline{1-1} \cline{4-6} 
Seq0007           &                     &                            & 677 (00:23)     & 162             & 25158          \\ \cline{1-1} \cline{4-6} 
Seq0008           &                     &                            & 577 (00:19)     & 133             & 10280          \\ \cline{1-1} \cline{4-6} 
Seq0009           &                     &                            & 526 (00:18)     & 193             & 27902          \\ \cline{1-1} \cline{4-6} 
Seq0010           &                     &                            & 569 (00:19)     & 101             & 22044          \\ \hline
\end{tabular}
\caption{Statistics of ant videos with annotations in indoor and outdoor scenes.}
\label{tab:Statistics}
\end{table*}

\subsection*{Data acquisition}

\subsubsection*{Indoor environment}

We collect 50 workers of Japanese arched ants species, which ranged from 7.4 to 13.8 $mm$ in body length~\cite{terayama1988two}.
We construct a laboratory environment for them, including a stable light source and a transparent plastic container.
We randomly divide them into 5 colonies of 10 ants each.
Then, we load each colony into the container in turns and film their activities with a high-resolution video camera.

\subsubsection*{Outdoor environment}
We collect 5 videos of black ant colonies in different outdoor environments, with the number of black ants ranging from 73 to 193 in the videos.
The body length of black ants is between 8 to 10 $mm$ ~\cite{ayieko2012nutritional}.
These videos are provided by Depositphotos, an online video site (\url{https://cn.depositphotos.com/home.html}), and access is subject to a royalty-free license.

\subsection*{Data Records}

The dataset consists of 10 image sequences from different scenes in JPEG digital image format, which is publicly available on \url{https://data.mendeley.com/datasets/9ws98g4npw/3}.
Alongside, we provide annotations marked by VisualMarkData for all image sequences in the form of text. 
In the dataset, the images and annotations of each sequence are organized into three folders are named 'det', 'gt', and 'img'.

\subsubsection*{Det folder}
The 'det' folder contains a 'det.txt' file which is the ground truth for detection, recording the location parameters of the ants in all frames, which is similar to multi-object tracking challenge~\cite{leal2015motchallenge}.
Each line represents one ant instance, and it contains 7 values as shown in Table~\ref{tab:data_format}.
The first number indicates in which frame the ant appears, while the second number identifies that ant as belonging to a trajectory by assigning a unique ID (set to -1 in a detection file, as no ID is assigned yet). 
The next four numbers indicate the location of the bounding box of the ant in 2D image coordinates. 
The location is indicated by the top-left corner as well as the width and height of the bounding box.
This is followed by a single number, which denotes the confidence score.

\subsubsection*{Gt folder}
The 'gt' folder contains a 'gt.txt' file, which is the ground truth for muti-object tracking. Similar to the 'det.txt' file, it also contains 7 values (see details in Table~\ref{tab:data_format}).
The difference compared to the 'det.txt' file is that the second number in the 'gt.txt' file records the ID of an ant as belonging to a trajectory, which provides the key information for implementing multi-ant tracking. 
Besides, each ant can be assigned to only one trajectory.

\subsubsection*{Img folder}
The 'img' folder stores the original image sequence converted from the video.
All images are converted to JPEG and named sequentially to a 6-digit file name (e.g. 000001.jpg).

\begin{table*}[]
\begin{tabular}{c|l|l}
\hline
\multicolumn{1}{l|}{\textbf{Position}} & \textbf{Name}       & \textbf{Description}                                                                                                                                                                                             \\ \hline
1                                      & Frame number        & Indicate at which frame the object is present                                                                                                                                                                    \\
2                                      & Identity number     & Each ant trajectory is identified by a unique ID (-1 for detections)                                                                                                                                      \\
3                                      & Bounding box left   & Coordinate of the top-left corner of the ant bounding box                                                                                                                                                 \\
4                                      & Bounding box top    & Coordinate of the top-left corner of the ant bounding box                                                                                                                                                 \\
5                                      & Bounding box width  & Width in pixels of the ant bounding box                                                                                                                                                                   \\
6                                      & Bounding box height & Height in pixels of the ant bounding box                                                                                                                                                                  \\
7                                      & Confidence score    & \begin{tabular}[c]{@{}l@{}}Indicates how confident the detector is that this instance is a ant. \\ For the ground truth and results, it acts as a flag whether the entry is to be considered\end{tabular} \\ \hline
\end{tabular}
\caption{Data format for annotation files, both for 'det.txt' and 'gt.txt' files.}
\label{tab:data_format}
\end{table*}


\section{Analyses}

\begin{figure*}[bht]
\centering
\includegraphics[width=\linewidth]{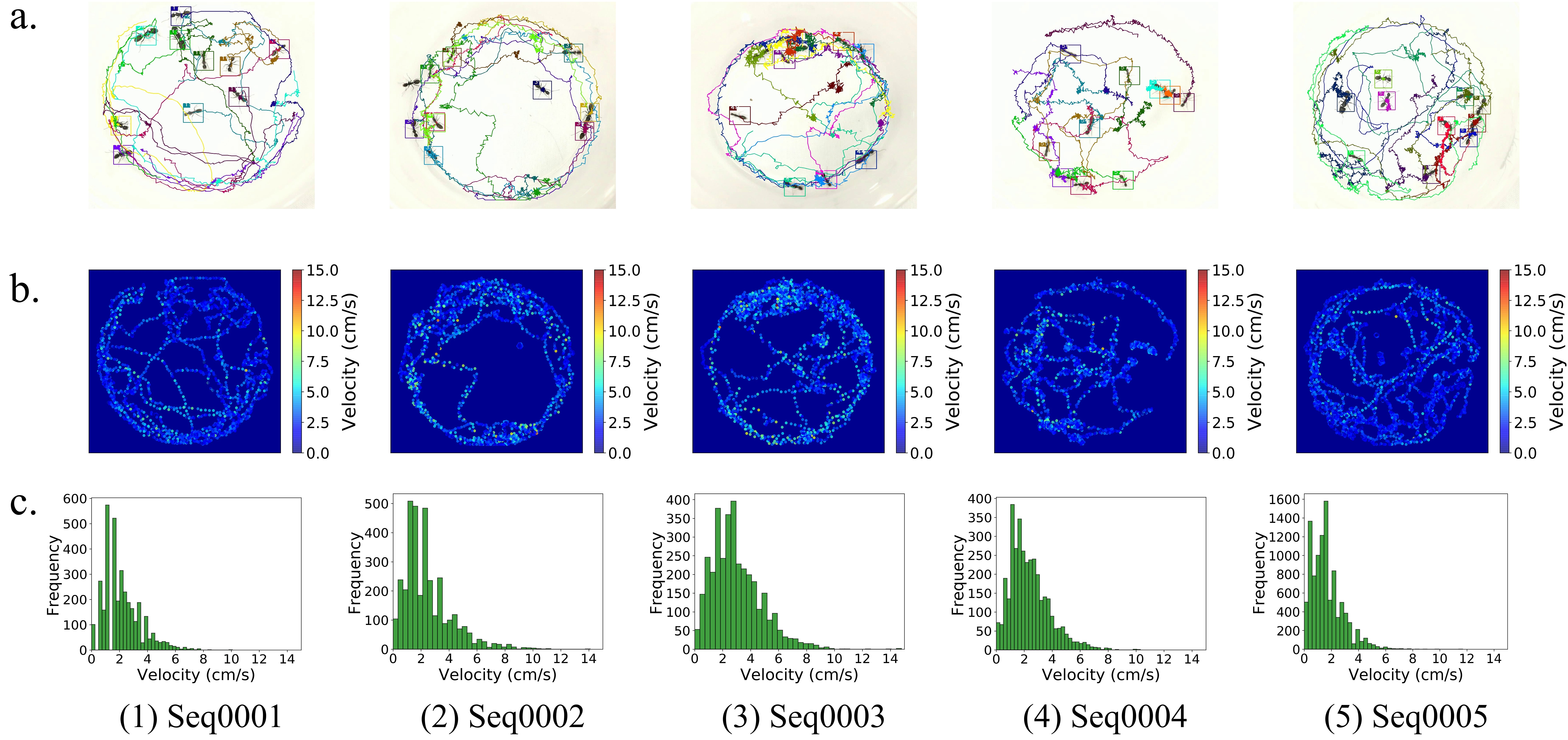}
\caption{Visual analysis of the marking results on indoor ant videos.
(a) Visualization of motion trajectories of the ants for each sequence of the indoor scene.
(b) Speed distributions in image space for 5 sequences of indoor scenes, respectively. 
(c) Ant speed histograms per indoor sequence.}
\label{fig:tech_val_in}
\end{figure*}

\begin{figure*}[bht]
\centering
\includegraphics[width=\linewidth]{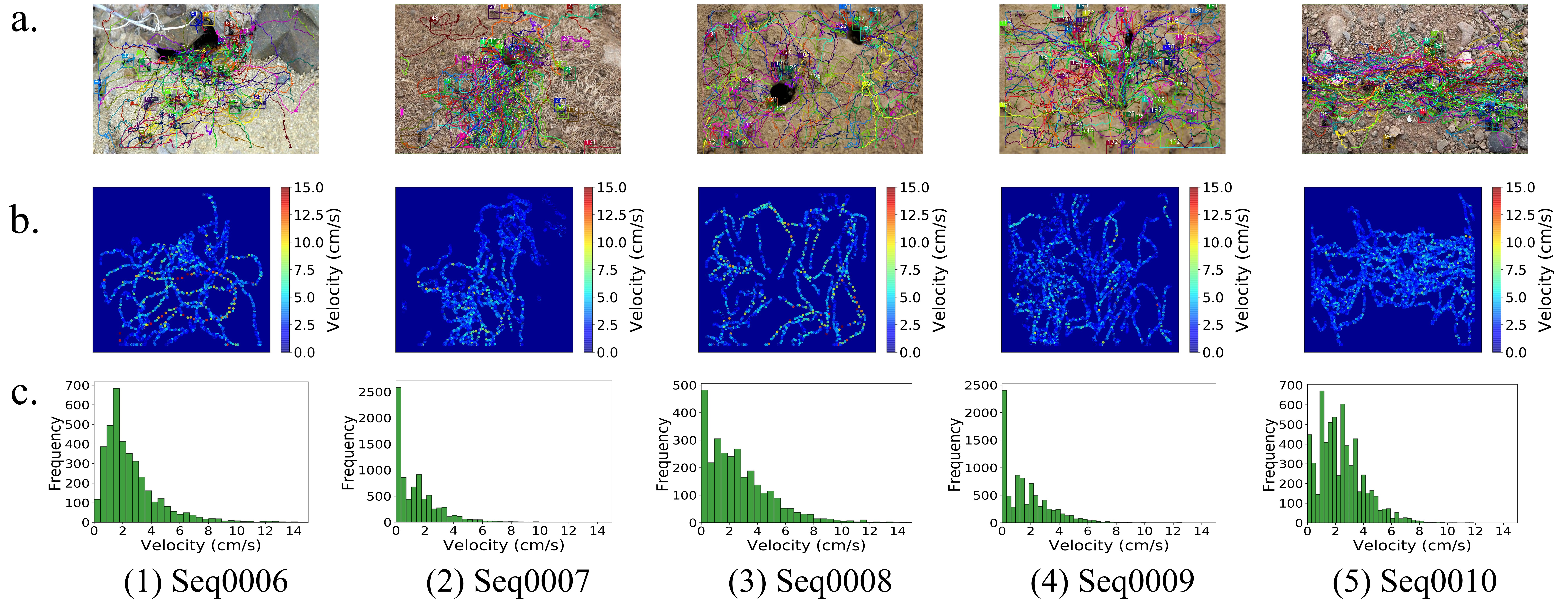}
\caption{Visual analysis of the marking results on outdoor ant videos.
(a) Visualization of motion trajectories of the ants for each sequence of the outdoor scene.
(b) Speed distributions in image space for 5 sequences of outdoor scenes, respectively. 
(c) Ant speed histograms per outdoor sequence.}
\label{fig:tech_val_out}
\end{figure*}

\subsection*{Visually confirm}

The ground truth annotations for all image sequences in the dataset are visually confirmed by the data annotation staff. 
The visual reviewing consists of two aspects, sequence-level (coarse-grained) and image-level (fine-grained).

Firstly, staffs perform a coarse-grained review of a single sequence.
Specifically, we draw the annotations on the corresponding images, and then convert the image sequence to video.
By replaying the video (see movies in \url{https://drive.google.com/file/d/1ibXbBhbCN8Ald_8SGV-luRBYUizZf-Zl/view?usp=sharing}), staff can quickly confirm which segments of the video are poor quality and needed to be re-marked.
For each scene, an example image frame is shown in Figure~\ref{fig:tech_val_in} (a) and Figure~\ref{fig:tech_val_out} (a).
After that, staff reviews the quality of annotations frame-by-frame via VisualMarkData.
For inaccurate annotations, staff modifies manually by using the "Check and modify" function of VisualMarkData (see details in Methods).

\subsection*{Motion speed analysis}

Further, to demonstrate the reliability of our dataset, we analyze the distribution of the movement speed of the ants in our dataset. 
First, for each ant, we use the 2D Euclidean distance~\cite{fabbri20082d} to calculate its pixel distance between two adjacent frames. 
Therefore, the pixel distance $\triangle ps_{t}$ of the ant at frame $t$ can be defined by the following equation:
\begin{equation}
\triangle ps_{t}=\sqrt{{(px_{t} - px_{t-1})^2}+{(py_{t} - py_{t-1})^2}}
\label{equ:disp}
\end{equation}

where $px_{t}$ denotes the pixel position of the ant in the horizontal direction at frame $t$. 
Similarly, $py_{t}$ denotes the pixel position in the vertical direction.
To convert the pixel distance to real-world coordinates, we divide the ant's body length $L$ (unit: $m$) in the real world by body length $n$ (unit: $pixel$) in the image.
Thus, the real-world displacement of the ant at frame $t$, $\triangle s_{t}$ (unit: $m$) can be expressed as follows:
\begin{equation}
\triangle s_{t} = \triangle ps_{t} \times L / n
\label{equ:st}
\end{equation}

Since the FPS for a specifc video is a constant $f_{c}$, the velocity $v_{t}$ (unit: $m \cdot s^{-1}$) at frame $t$ can be formulated as:
\begin{equation}
v_{t} = \frac{\triangle s_{t}}{1/f_{c}}
\label{equ:vt}
\end{equation}
Where, the $v_{0}$ is set to 0.

According to the aforementioned equations, combined with the location information of ants in annotations, we can analyze the motion speed of ants in the video, as shown in Figure~\ref{fig:tech_val_in} (b), (c) and Figure~\ref{fig:tech_val_out} (b), (c).
Specifically, the overall motion speed of ants in indoor and outdoor scenes are 2.16$\pm$1.49 $cm \cdot s^{-1}$ and 1.98$\pm$1.84 $cm \cdot s^{-1}$, respectively.
These values are within a reasonable range (ants average motion speed is 2.85 $cm \cdot s^{-1}$ under bi-directional traffic condition~\cite{wang2018bi}).
This demonstrates that the ant colony activity dataset we collected and marked is real and reliable.

\section{Discussion}


The image sequence marking software VisualMarkData is a toolkit with interactive visualization. The goal of the software is to provide a convenient tool for researchers marking movement trajectories of social insects in videos, thus facilitating the study of the behavioral mechanisms of social insects.
Additionally, by using the software, researchers will obtain standardized annotation data, as details in the previous section.
VisualMarkData is open source, which enables researchers to mark their image sequence datasets of any multi-object motion scenario.
Alongside, we have provided publicly available Python Scripts to illustrate the analysis of data as well as usage of the data.
To visualize and reproduce the results described in the Technical Validation section, we develop two scripts for the researchers.
Also, we provide another script to calculate metrics~\cite{leal2015motchallenge} of multi-object tracking that enables any deep learning algorithm to evaluate the tracking accuracy on our dataset.
The annotated trajectory data can be used for training and testing of supervised learning models, thus providing a powerful tool for studying a wider range of ant colony behaviors.

In the future, we will enrich the VisualMarkData with more features to reduce the difficulty of marking and improve the efficiency of marking. The software currently marks targets based on their center points, and we are considering introducing stretchable annotation capabilities based on rectangles or ellipses.
In addition, the simultaneous annotation of multiple targets in one frame is also a feature worth developing. Along with that, we can introduce semi-automated annotation, i.e., embedding a neural network model into the VisualMarkData, which will automatically predict and annotate objects of the current frame based on the information in the previous frame. Thus, the annotators only need to fine-tune the annotation, which will significantly improve the efficiency of the annotation.

The dataset and VisualMarkData will boost researchers both in biology and computer science to study on behavior of social insects in different environments. We hope that this work will contribute to the potential discovery of ant colony behavioral mechanisms and facilitate the application of the image processing field in biology.

\section{Potential implications}



Swarm behavior is one of the most important features of social insects, which has important significance for the study of embodied intelligence ~\cite{tiacharoen2012design}. Specifically, social insects often tend to cluster into a colony~\cite{vandermeer2008clusters}, which forms a complex dynamical system together with the surrounding environment~\cite{balch2001automatically}.
So far, researchers do not know enough about the mechanisms behind swarm behaviors of social insects.
We believe our image sequence marking software and dataset could facilitate the analysis of ant colony behavior leading to the development of embodied intelligence.

\section{Methods}

\subsection*{Hardware devices for acquiring raw data}

For indoor environments, we use a cylindrical container made of transparent plastic providing a space for the ants to move around. 
This container has a bottom diameter of 10 $cm$, a side height of 15 $cm$, and is not closed at the top.
Ants, loaded in the container, are filmed with a high-resolution video camera (Panasonic GX 85) with 25 frames per second (FPS) in the format H.264 with a resolution of 1920$\times$1080 pixels. 
And the distance between the camera and the top of the container is 30 $cm$ so that the filming view of the camera can cover the whole container.
To ensure stable filming, we fix the camera on a tripod, as well as hanging a light bulb above the container.
Besides, the anti-dusting powder is applied to the inner wall of the container, preventing ants from escaping from the container during the filming.

\subsection*{Description of the marking software VisualMarkData}

\begin{figure*}[ht]
\centering
\includegraphics[width=\linewidth]{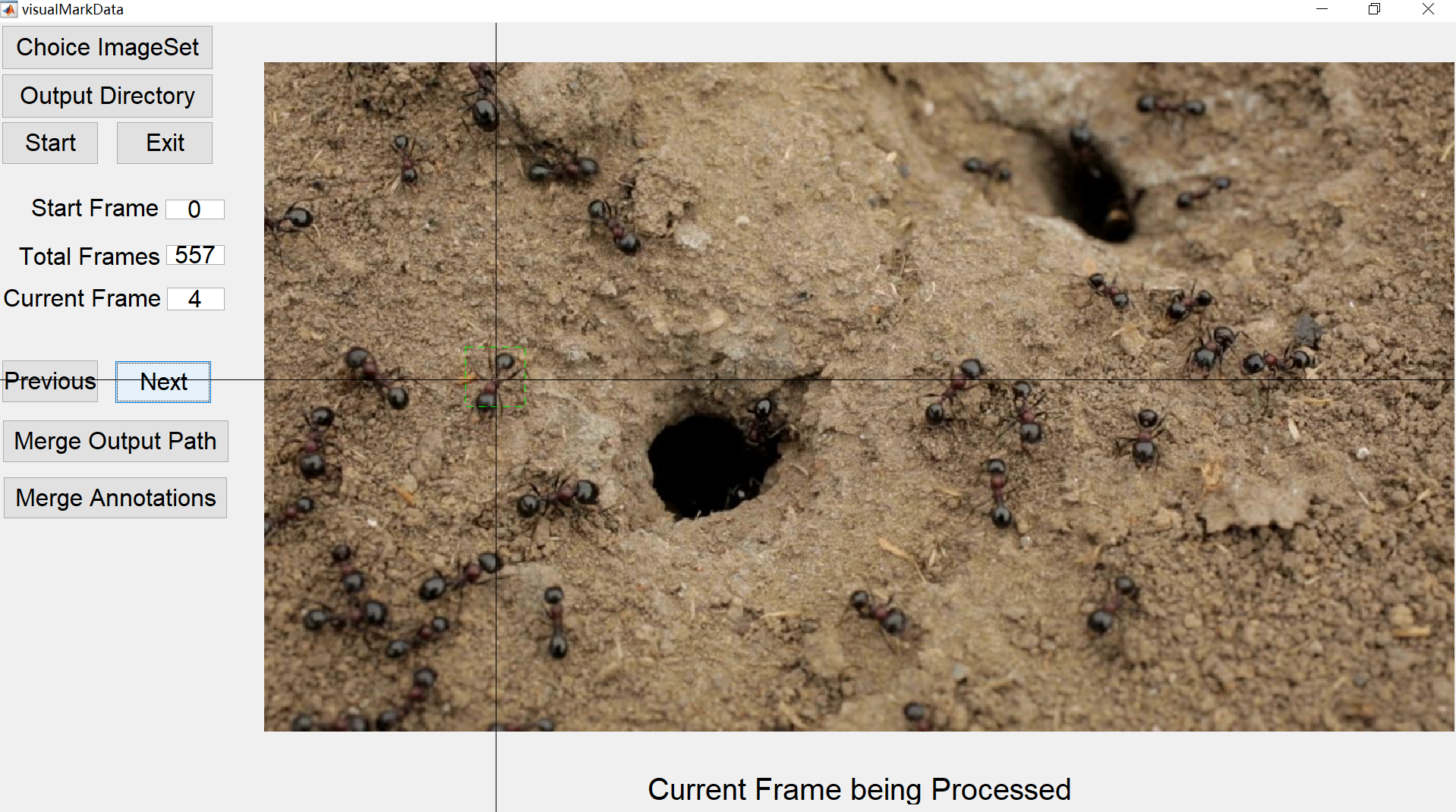}
\caption{The interactive interface of our image sequence marking software is named VisualMarkData.
After selecting an image sequence, the user can acquire the annotation by clicking on the center location of the ant's body.}
\label{fig:VisualMarkData}
\end{figure*}

We develop an image sequence marking software called VisualMarkData to provide the locations and identification numbers of objects in the sequence for motion analysis.
The operation procedure of VisualMarkData is as follows, and its interface is shown in Figure~\ref{fig:VisualMarkData}.
\begin{itemize}
    \item \textbf{Choose Image Set.}
    Before marking, the user should click "Choose ImageSet" to select an image set. 
    The filename of the image set is defined in the format of "SeqXObjectYImageZ", where X is the name of the sequence, Y is the number of objects in the first frame and Z is the size of the bounding box which represents the object.
    For example, the image set, named "Seq0001Object10Image94", indicates that the sequence "0001" contains 10 objects in the first frame, and each object will be marked with a bounding box with the size of 94x94.

    \item \textbf{Create Output Directory.} 
    The user needs to click "Output Directory" to select the storage path of annotations. 
    Since VisualMarkData only focuses on one object per marking round (each round goes through the whole image sequence), the output folder is suggested to be named with the identification number of the object, e.g. "0001". As the identity number of the object is user-defined, the user can use any number for the object and folder as long as it is unique.

    \item \textbf{Select Start Frame.}
    In the last step before starting marking, you need to enter the start frame, the default value is 0.
    This means that you are allowed to exit the software halfway and continue the progress of the current marking task the next time. Then, you can click the "Start" button.

    \item \textbf{Marking.}
    The user clicks on the center of the object in the current frame, and the software will automatically save the digital location of the center, as well as a bounding box centered on the object.
    It should be emphasized that the user only marks the same object until finishes the entire image sequence, and then the user can focus on another object by repeating the same operation for the previous one. 

    \item \textbf{Next Frame.}
    The user clicks the "Next" button to show the next frame on the window of the software.
    The marked location on the previous frame will be displayed with a green-dotted, which can help the user quickly locate the target object.

    \item \textbf{Previous Frame.}
    If the marked location of the previous frame is incorrect, the user can click the "Previous" button to roll back one frame.

    \item \textbf{Check and Modify.}
    After the user finishes marking the entire image set, checking is needed to guarantee the quality.
    In this case, the user can enter the specific frame to modify the annotations by carrying out \textbf{Select Start Frame} step.

    \item \textbf{Merge Annotations.}
    After all objects in a sequence have been marked and reviewed, the user needs to click the "Merge" button, thereby all annotations for each object will be sorted by frames and then the ID of the object both in ascending order.

\end{itemize}

\section{Availability of source code and requirements}

Lists the following:
\begin{itemize}
\item Project name: ANTS\_marking\_and\_analysis\_tools
\item Project home page: e.g.~\url{https://github.com/holmescao/ANTS_marking_and_analysis_tools}
\item Operating system(s): Platform independent
\item Programming language: Python, MATLAB, Shell
\item Other requirements: MATLAB R2021b (with Image Processing Toolbox)
\item License: MIT License

\end{itemize}


\section{Availability of supporting data and materials}






The data set supporting the results of this article is available in the ANTS--ant detection and tracking repository at \url{https://data.mendeley.com/datasets/9ws98g4npw/3}.

\section{Declarations}

\subsection{List of abbreviations}

CV: computer vision;
ID: identity;
FPS: frames per second







\subsection{Consent for publication}



Not applicable

\subsection{Competing Interests}


The authors declare that they have no competing interests.

\subsection{Funding}


This work was supported by the Natural Science Foundation of Fujian Province of China (No. 2019J01002).

\subsection{Author's Contributions}

M.W. and S.G conceived the experiment(s), X.C. conducted the experiment(s), X.C. analyzed the results. All authors reviewed the manuscript.

\section{Acknowledgements}





The authors thank the reviewers, for providing useful suggestions for improvements and valuable feedback on the workflow and the manuscript.



\bibliography{ref}

\begin{thebibliography}{31}
\providecommand{\natexlab}[1]{#1}
\providecommand{\url}[1]{\texttt{#1}}
\providecommand{\urlprefix}{}

\bibitem[{Vandermeer et~al.(2008)Vandermeer, John and Perfecto, Ivette and
  Philpott, Stacy M}]{vandermeer2008clusters}
Vandermeer J, Perfecto I, Philpott SM.
\newblock Clusters of ant colonies and robust criticality in a tropical
  agroecosystem.
\newblock Nature 2008;451(7177):457--459.

\bibitem[{Balch et~al.(2001)Balch, Tucker and Khan, Zia and Veloso,
  Manuela}]{balch2001automatically}
Balch T, Khan Z, Veloso M.
\newblock Automatically tracking and analyzing the behavior of live insect
  colonies.
\newblock In: Proceedings of the fifth international conference on Autonomous
  agents; 2001. p. 521--528.

\bibitem[{H{\"o}lldobler et~al.(1990)H{\"o}lldobler, Bert and Wilson, Edward O
  and others}]{holldobler1990ants}
H{\"o}lldobler B, Wilson EO, et~al.
\newblock The ants.
\newblock Harvard University Press; 1990.

\bibitem[{Whitehouse and Jaffe(1996)Whitehouse, Mary EA and Jaffe,
  Klaus}]{whitehouse1996ant}
Whitehouse ME, Jaffe K.
\newblock Ant wars: combat strategies, territory and nest defence in the
  leaf-cutting antAtta laevigata.
\newblock Animal Behaviour 1996;51(6):1207--1217.

\bibitem[{Vaughan et~al.(2000)Vaughan, Richard T and St{\o}y, Kasper and
  Sukhatme, Gaurav S and Matari{\'c}, Maja J}]{vaughan2000whistling}
Vaughan RT, St{\o}y K, Sukhatme GS, Matari{\'c} MJ.
\newblock Whistling in the dark: cooperative trail following in uncertain
  localization space.
\newblock In: Proceedings of the fourth international conference on Autonomous
  agents; 2000. p. 187--194.

\bibitem[{Fewell(2003)Fewell, Jennifer H}]{fewell2003social}
Fewell JH.
\newblock Social insect networks.
\newblock Science 2003;301(5641):1867--1870.

\bibitem[{Motani et~al.(2005)Motani, Mehul and Srinivasan, Vikram and
  Nuggehalli, Pavan S}]{motani2005peoplenet}
Motani M, Srinivasan V, Nuggehalli PS.
\newblock Peoplenet: engineering a wireless virtual social network.
\newblock In: Proceedings of the 11th annual international conference on Mobile
  computing and networking; 2005. p. 243--257.

\bibitem[{Tiacharoen and Chatchanayuenyong(2012)Tiacharoen, S and
  Chatchanayuenyong, T}]{tiacharoen2012design}
Tiacharoen S, Chatchanayuenyong T.
\newblock Design and development of an intelligent control by using bee colony
  optimization technique.
\newblock American Journal of Applied Sciences 2012;9(9):1464.

\bibitem[{Poff et~al.(2012)Poff, Corey and Nguyen, Hoan and Kang, Timothy and
  Shin, Min C}]{poff2012efficient}
Poff C, Nguyen H, Kang T, Shin MC.
\newblock Efficient tracking of ants in long video with GPU and interaction.
\newblock In: 2012 IEEE Workshop on the Applications of Computer Vision (WACV)
  IEEE; 2012. p. 57--62.

\bibitem[{Khan et~al.(2005)Khan, Zia and Balch, Tucker and Dellaert,
  Frank}]{khan2005mcmc}
Khan Z, Balch T, Dellaert F.
\newblock MCMC-based particle filtering for tracking a variable number of
  interacting targets.
\newblock IEEE transactions on pattern analysis and machine intelligence
  2005;27(11):1805--1819.

\bibitem[{Khan et~al.(2006)Khan, Zia and Balch, Tucker and Dellaert,
  Frank}]{khan2006mcmc}
Khan Z, Balch T, Dellaert F.
\newblock MCMC data association and sparse factorization updating for real time
  multitarget tracking with merged and multiple measurements.
\newblock IEEE transactions on pattern analysis and machine intelligence
  2006;28(12):1960--1972.

\bibitem[{Oh et~al.(2006)Oh, Sang Min and Rehg, James M and Dellaert,
  Frank}]{oh2006parameterized}
Oh SM, Rehg JM, Dellaert F.
\newblock Parameterized duration mmodeling for switching linear dynamic
  systems.
\newblock In: 2006 IEEE Computer Society Conference on Computer Vision and
  Pattern Recognition (CVPR'06), vol.~2 IEEE; 2006. p. 1694--1700.

\bibitem[{Veeraraghavan et~al.(2008)Veeraraghavan, Ashok and Chellappa, Rama
  and Srinivasan, Mandyam}]{veeraraghavan2008shape}
Veeraraghavan A, Chellappa R, Srinivasan M.
\newblock Shape-and-behavior encoded tracking of bee dances.
\newblock IEEE transactions on pattern analysis and machine intelligence
  2008;30(3):463--476.

\bibitem[{Fletcher et~al.(2011)Fletcher, Mary and Dornhaus, Anna and Shin, Min
  C}]{fletcher2011multiple}
Fletcher M, Dornhaus A, Shin MC.
\newblock Multiple ant tracking with global foreground maximization and
  variable target proposal distribution.
\newblock In: 2011 IEEE Workshop on Applications of Computer Vision (WACV)
  IEEE; 2011. p. 570--576.

\bibitem[{Li et~al.(2008)Li, Min and Zhang, Zhaoxiang and Huang, Kaiqi and Tan,
  Tieniu}]{li2008estimating}
Li M, Zhang Z, Huang K, Tan T.
\newblock Estimating the number of people in crowded scenes by mid based
  foreground segmentation and head-shoulder detection.
\newblock In: 2008 19th international conference on pattern recognition IEEE;
  2008. p. 1--4.

\bibitem[{Li et~al.(2009)Li, Yuan and Huang, Chang and Nevatia,
  Ram}]{li2009learning}
Li Y, Huang C, Nevatia R.
\newblock Learning to associate: Hybridboosted multi-target tracker for crowded
  scene.
\newblock In: 2009 IEEE conference on computer vision and pattern recognition
  IEEE; 2009. p. 2953--2960.

\bibitem[{Zhao et~al.(2015)Zhao, Miaomiao and Liu, Hongxia and Wan,
  Yi}]{zhao2015improved}
Zhao M, Liu H, Wan Y.
\newblock An improved Canny edge detection algorithm based on DCT.
\newblock In: 2015 IEEE International Conference on Progress in Informatics and
  Computing (PIC) IEEE; 2015. p. 234--237.

\bibitem[{Schmelzer and Kastberger(2009)Schmelzer, Evelyn and Kastberger,
  Gerald}]{schmelzer2009special}
Schmelzer E, Kastberger G.
\newblock ‘Special agents’ trigger social waves in giant honeybees (Apis
  dorsata).
\newblock Naturwissenschaften 2009;96(12):1431--1441.

\bibitem[{Kastberger et~al.(2013)Kastberger, Gerald and Weihmann, Frank and
  Hoetzl, Thomas}]{kastberger2013social}
Kastberger G, Weihmann F, Hoetzl T.
\newblock Social waves in giant honeybees (Apis dorsata) elicit nest
  vibrations.
\newblock Naturwissenschaften 2013;100(7):595--609.

\bibitem[{Tan et~al.(2016)Tan, Ken and Dong, Shihao and Li, Xinyu and Liu,
  Xiwen and Wang, Chao and Li, Jianjun and Nieh, James C}]{tan2016honey}
Tan K, Dong S, Li X, Liu X, Wang C, Li J, et~al.
\newblock Honey bee inhibitory signaling is tuned to threat severity and can
  act as a colony alarm signal.
\newblock PLoS biology 2016;14(3):e1002423.

\bibitem[{Dong et~al.(2018)Dong, Shihao and Wen, Ping and Zhang, Qi and Wang,
  Yuan and Cheng, Yanan and Tan, Ken and Nieh, James C}]{dong2018olfactory}
Dong S, Wen P, Zhang Q, Wang Y, Cheng Y, Tan K, et~al.
\newblock Olfactory eavesdropping of predator alarm pheromone by sympatric but
  not allopatric prey.
\newblock Animal Behaviour 2018;141:115--125.

\bibitem[{Schor et~al.(2016)Schor, Noa and Bechar, Avital and Ignat, Timea and
  Dombrovsky, Aviv and Elad, Yigal and Berman, Sigal}]{schor2016robotic}
Schor N, Bechar A, Ignat T, Dombrovsky A, Elad Y, Berman S.
\newblock Robotic disease detection in greenhouses: Combined detection of
  powdery mildew and tomato spotted wilt virus.
\newblock IEEE Robotics and Automation Letters 2016;1(1):354--360.

\bibitem[{Wang et~al.(2018)Wang, Guotai and Li, Wenqi and Zuluaga, Maria A and
  Pratt, Rosalind and Patel, Premal A and Aertsen, Michael and Doel, Tom and
  David, Anna L and Deprest, Jan and Ourselin, S{\'e}bastien and
  others}]{wang2018interactive}
Wang G, Li W, Zuluaga MA, Pratt R, Patel PA, Aertsen M, et~al.
\newblock Interactive medical image segmentation using deep learning with
  image-specific fine tuning.
\newblock IEEE transactions on medical imaging 2018;37(7):1562--1573.

\bibitem[{Wang(2018)Wang, Canyong}]{wang2018research}
Wang C.
\newblock Research and application of traffic sign detection and recognition
  based on deep learning.
\newblock In: 2018 International Conference on Robots \& Intelligent System
  (ICRIS) IEEE; 2018. p. 150--152.

\bibitem[{Imirzian et~al.(2019)Imirzian, Natalie and Zhang, Yizhe and Kurze,
  Christoph and Loreto, Raquel G and Chen, Danny Z and Hughes, David
  P}]{imirzian2019automated}
Imirzian N, Zhang Y, Kurze C, Loreto RG, Chen DZ, Hughes DP.
\newblock Automated tracking and analysis of ant trajectories shows variation
  in forager exploration.
\newblock Scientific reports 2019;9(1):1--10.

\bibitem[{Cao et~al.(2020)Cao, Xiaoyan and Guo, Shihui and Lin, Juncong and
  Zhang, Wenshu and Liao, Minghong}]{cao2020online}
Cao X, Guo S, Lin J, Zhang W, Liao M.
\newblock Online tracking of ants based on deep association metrics: method,
  dataset and evaluation.
\newblock Pattern Recognition 2020;103:107233.

\bibitem[{Terayama and Ogata(1988)Terayama, Mamoru and Ogata,
  Kazuo}]{terayama1988two}
Terayama M, Ogata K.
\newblock Two new species of the ant genus Probolomyrmex (Hymenoptera,
  Formicidae) from Japan.
\newblock Kontyu 1988;56(3):590--594.

\bibitem[{Ayieko et~al.(2012)Ayieko, Monica A and Kinyuru, JN and Ndong’a, MF
  and Kenji, GM}]{ayieko2012nutritional}
Ayieko MA, Kinyuru J, Ndong’a M, Kenji G.
\newblock Nutritional value and consumption of black ants (Carebara vidua
  Smith) from the Lake Victoria region in Kenya.
\newblock Advance Journal of Food Science and Technology 2012;.

\bibitem[{Leal-Taix{\'e} et~al.(2015)Leal-Taix{\'e}, Laura and Milan, Anton and
  Reid, Ian and Roth, Stefan and Schindler, Konrad}]{leal2015motchallenge}
Leal-Taix{\'e} L, Milan A, Reid I, Roth S, Schindler K.
\newblock Motchallenge 2015: Towards a benchmark for multi-target tracking.
\newblock arXiv preprint arXiv:150401942 2015;.

\bibitem[{Fabbri et~al.(2008)Fabbri, Ricardo and Costa, Luciano Da F and
  Torelli, Julio C and Bruno, Odemir M}]{fabbri20082d}
Fabbri R, Costa LDF, Torelli JC, Bruno OM.
\newblock 2D Euclidean distance transform algorithms: A comparative survey.
\newblock ACM Computing Surveys (CSUR) 2008;40(1):1--44.

\bibitem[{Wang et~al.(2018)Wang, Qiao and Song, Weiguo and Zhang, Jun and Lo,
  Siuming}]{wang2018bi}
Wang Q, Song W, Zhang J, Lo S.
\newblock Bi-directional movement characteristics of Camponotus japonicus ants
  during nest relocation.
\newblock Journal of Experimental Biology 2018;221(18):jeb181669.

\end{thebibliography}

\end{document}